\def\year{2020}\relax
\documentclass[letterpaper]{article} 
\usepackage{aaai20}  
\usepackage{times}  
\usepackage{helvet} 
\usepackage{courier}  
\usepackage[hyphens]{url}  
\usepackage{graphicx} 
\usepackage{graphicx}  
\usepackage{graphicx}   
\usepackage{subfigure}
\usepackage{multirow}
\usepackage{diagbox}
\usepackage{color}		
\usepackage{bbm}
\usepackage{multicol}
\usepackage{blindtext}
\usepackage{amsmath,amsfonts,amssymb}
\usepackage{bm}
\usepackage{mathtools}
\usepackage{wrapfig}
\usepackage{wrapfig,lipsum,booktabs}
\usepackage{algorithm}
\usepackage{algpseudocode}

\DeclareMathOperator*{\softmax}{softmax}

\DeclareMathOperator*{\sigmoid}{sigmoid}
\DeclareMathOperator*{\onedcnn}{1D-CNN}

\DeclareMathOperator*{\pool}{Pool}
\DeclareMathOperator*{\mlp}{MLP}

\title{Self-Attention Enhanced Selective Gate with Entity-Aware Embedding for Distantly Supervised Relation Extraction}
\author{
	Yang Li\textsuperscript{1},
	Guodong Long \thanks{~Corresponding author}\textsuperscript{1},
	Tao Shen\textsuperscript{1},
	Tianyi Zhou\textsuperscript{2},
	Lina Yao\textsuperscript{3},
	Huan Huo\textsuperscript{1},
	Jing Jiang\textsuperscript{1} \\
	\textsuperscript{1} Centre of Artificial Intelligence, FEIT, University of Technology Sydney\\
	\textsuperscript{2} Paul G. Allen School of Computer Science \& Engineering, University of Washington\\
	\textsuperscript{3} School of Computer Science and Engineering, University of New South Wales\\
	\texttt{\{yang.li-17, tao.shen\}@student.uts.edu.au} \\
    \texttt{tianyizh@uw.edu, lina.yao@unsw.edu.au} \\
	\texttt{\{guodong.long, huan.huo, jing.jiang\}@uts.edu.au} \\
}

\begin{document}

\maketitle

\begin{abstract}
Distantly supervised relation extraction intrinsically suffers from noisy labels due to the strong assumption of distant supervision. Most prior works adopt a selective attention mechanism over sentences in a bag to denoise from wrongly labeled data, which however could be incompetent when there is only one sentence in a bag. In this paper, we propose a brand-new light-weight neural framework to address the distantly supervised relation extraction problem and alleviate the defects in previous selective attention framework. Specifically, in the proposed framework, 1) we use an entity-aware word embedding method to integrate both relative position information and head/tail entity embeddings, aiming to highlight the essence of entities for this task; 2) we develop a self-attention mechanism to capture the rich contextual dependencies as a complement for local dependencies captured by piecewise CNN; and 3) instead of using selective attention, we design a pooling-equipped gate, which is based on rich contextual representations, as an aggregator to generate bag-level representation for final relation classification. Compared to selective attention, one major advantage of the proposed gating mechanism is that, it performs stably and promisingly even if only one sentence appears in a bag and thus keeps the consistency across all training examples. The experiments on NYT dataset demonstrate that our approach achieves a new state-of-the-art performance in terms of both AUC and top-n precision metrics.  
\end{abstract}

\section{Introduction}
Relation extraction (RE) is one of the most fundamental tasks in natural language processing, and its goal is to identify the relationship between a given pair of entities in a sentence. 
Typically, a large-scale training dataset with clean labels is required to train a reliable relation extraction model. However, it is time-consuming and labor-intensive to annotate such data by crowdsourcing. 
To overcome the lack of labeled training data, \citeauthor{mintz2009distant}~\shortcite{mintz2009distant} presents a distant supervision approach that automatically generates a large-scale, labeled training set by aligning entities in knowledge graph (e.g. Freebase \cite{bollacker2008freebase}) to corresponding entity mentions in natural language sentences. 
This approach is based on a \emph{strong assumption} that, any sentence containing two entities should be labeled according to the relationship of the two entities on the given knowledge graph. However, this assumption does not always hold. 
Sometimes the same two entities in different sentences with various contexts cannot express a consistent relationship as described in the knowledge graph, which certainly results in wrongly labeled problem. 

To alleviate the aformentioned problem, \citeauthor{riedel2010modeling}~\shortcite{riedel2010modeling} proposes a multi-instance learning framework, which relaxes the strong assumption to \emph{expressed-at-least-one} assumption.
In plainer terms, this means any possible relation between two entities hold true in at least one distantly-labeled sentence rather than all of the them that contains those two entities. 
In particular, instead of generating a sentence-level label, this framework assigns a label to a \emph{bag} of sentences containing a common entity pair, and the label is a relationship of the entity pair on knowledge graph. 
Recently, based on the labeled data at bag level, a line of works \cite{zeng2015distant,du2018multi,lin2016neural,han2018hierarchical,ye2019distant} under selective attention framework \cite{lin2016neural} let model implicitly focus on the correctly labeled sentence(s) by an attention mechanism and thus learn a stable and robust model from the noisy data.

\begin{table}[t] \small
	\centering
	\begin{tabular}{p{4.5cm}p{1.5cm}p{1cm}}
		\toprule
		\textbf{Bag consisting of one sentence}&\textbf{Label}&\textbf{Correct}  \\
		\midrule
		After moving back to \emph{New York}, \emph{Miriam} was the victim of a seemingly racially motivated attack ... &  place\_lived& True\\ \midrule
		... he faced, walking \emph{Bill Mueller} and giving up singles to Mark Bellhorn and \emph{Johnny Damon}. &  place\_lived& False\\
		\bottomrule
	\end{tabular}
	\caption{\small Two examples of one-sentence bag, which are correctly and wrongly labeled by distant supervision respectively.}
	\label{tab:nosiylabel}
\end{table}

However, such selective attention framework is vulnerable to situations where a bag is merely comprised of one single sentence labeled;  
and what is worse, the only one sentence possibly expresses inconsistent relation information with the bag-level label. 
This scenario is not uncommon. For a popular distantly supervised relation extraction benchmark, e.g., NYT dataset \cite{riedel2010modeling}, up to $80\%$ of its training examples (i.e., bags) are one-sentence bags. From our data inspection, we randomly sample 100 one-sentence bags and find $35\%$ of them is incorrectly labeled. Two examples of one-sentence bag are shown in Table \ref{tab:nosiylabel}. 
These results indicate that, in training phrase the selective attention module is enforced to output a single-valued scalar for $80\%$ examples, leading to an ill-trained attention module and thus hurting the performance. 

Motivated by aforementioned observations, in this paper, we propose a novel \textbf{Se}lective \textbf{G}ate (SeG) framework for distantly supervised relation extraction. 
In the proposed framework, 
1) we employ both the entity embeddings and relative position embeddings \cite{zeng2014relation} for relation extraction, and an entity-aware embedding approach is proposed to dynamically integrate entity information into each word embedding, yielding more expressively-powerful representations for downstream modules; 
2) to strengthen the capability of widely-used piecewise CNN (PCNN) \cite{zeng2015distant} on capturing long-term dependency \cite{yu2018qanet}, we develop a light-weight self-attention \cite{lin2017structured,shen2017disan} mechanism to capture rich dependency information and consequently enhance the capability of neural network via producing complementary representation for PCNN; 
and 3) based on preceding versatile features, we design a selective gate to aggregate sentence-level representations into bag-level one and alleviate intrinsic issues appearing in selective attention. 

Compared to the baseline framework (i.e., selective attention for multi-instance learning), SeG is able to produce entity-aware embeddings and rich-contextual representations to facilitate downstream aggregation modules that stably learn from noisy training data. 
Moreover, SeG uses gate mechanism with pooling to overcome problem occurring in selective attention, which is caused by one-sentence bags. In addition, it still keeps a light-weight structure to ensure the scalability of this model.

The experiments and extensive ablation studies on New York Time dataset \cite{riedel2010modeling} show that our proposed framework achieves a new state-of-the-art performance regarding both AUC and top-n precision metrics for distantly supervised relation extraction task, and also verify the significance of each proposed module. Particularly, the proposed framework can achieve AUC of 0.51, which outperforms selective attention baseline by 0.14 and improves previous state-of-the-art approach by 0.09.

\section{Proposed Approach}

\begin{figure*}[t]
    \centering
    \includegraphics[width=0.75\textwidth]{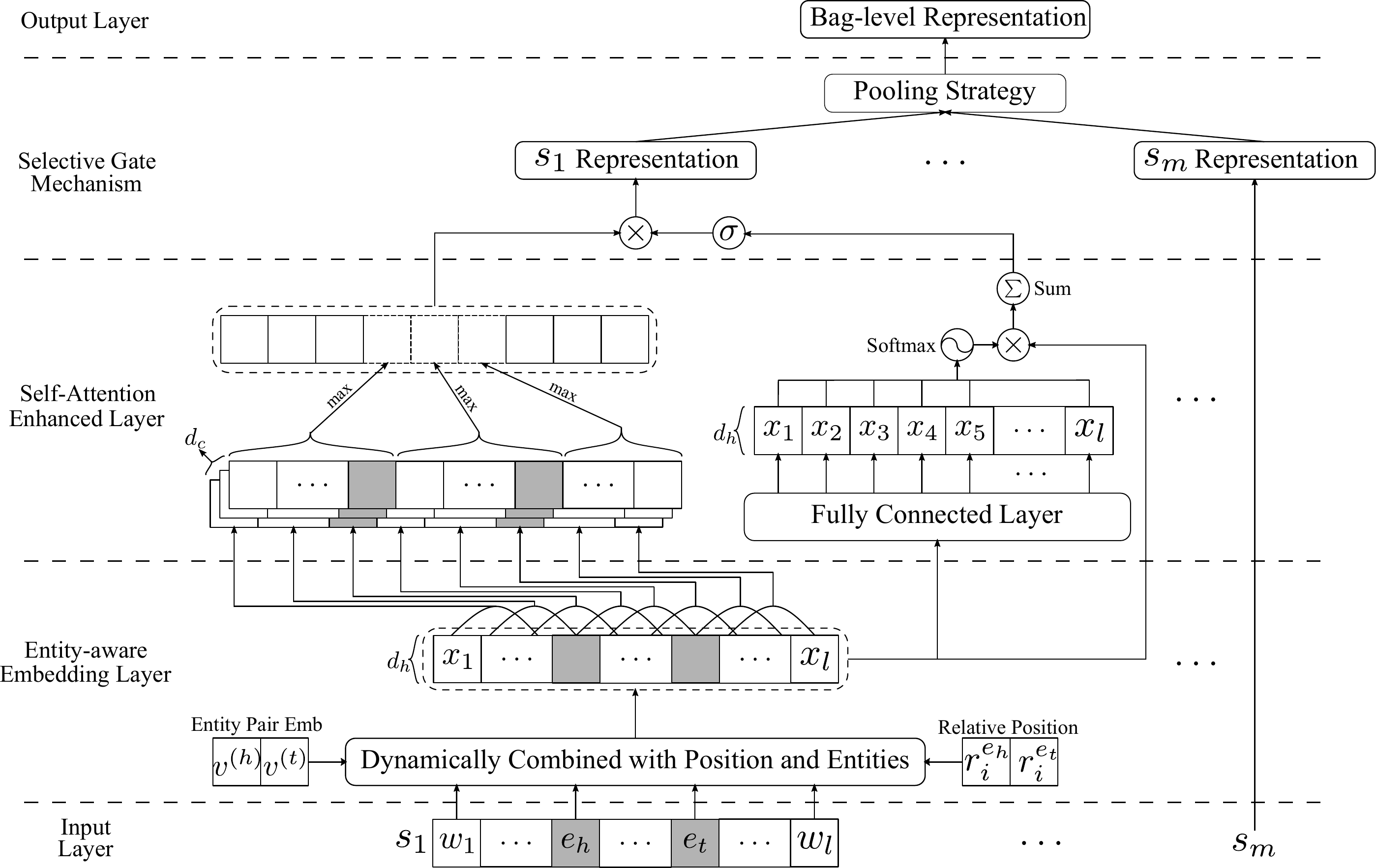}
    \caption{\small The framework of our approach (i.e. SeG) that consisting of three components: 1) entity-aware embedding 2) self-attention enhanced neural network and 3) a selective gate. Note, tokens $e_h$ and $e_t$ with gray background mean the head entity and tail entity of this sentence.}
    \label{fig:model}
\end{figure*}

As illustrated in Figure \ref{fig:model}, we propose a novel neural network, i.e., SeG, for distantly supervised relation extraction, which is composed of following neural components. 

\subsection{Entity-Aware Embedding} \label{sec:ent_aware}

Given a bag of sentences\footnote{``sentence'' and ``instance'' are interchangeable in this paper.} $B^k = \{s^k_1, \dots, s^k_{m^k}\}$ where each sentence contains common entity pair (i.e., head entity $e^k_h,$ and tail entity $e^k_t$), the target of relation extraction is to predict the relation $y^k$ between the two entities. For a clear demonstration, we omit indices of example and sentence in remainder if no confusion caused. Each sentence is a sequence of tokens, i.e., $s = [w_1, \dots, w_n]$, where $n$ is the length of the sentence. In addition, each token has a low-dimensional dense-vector representation, i.e., $[\bm{v}_1, \cdots, \bm{v}_n] \in\mathbb R^{d_w \times n}$, where $d_w$ denotes the dimension of word embedding. 

In addition to the typical word embedding, relative position is a crucial feature for relation extraction, which can provide downstream neural model with rich positional information \cite{zeng2014relation,zeng2015distant}. Relative positions explicitly describe the relative distances between each word $w_i$ and the two targeted entities $e_h$ and $e_t$. For $i$-th word, a randomly initialized weight matrix projects the relative position features into a two dense-vector representations w.r.t the head and tail entities, i.e., $\bm{r}^{e_h}_i$ and $\bm{r}^{e_t}_i\in\mathbb R^{d_r}$ respectively. The final low-level representations for all tokens are a concatenation of the aforementioned embeddings, i.e., $\bm{X}^{(p)} = [\bm{x}^{(p)}_1, \cdots, \bm{x}^{(p)}_n] \in\mathbb R^{d_p \times n}$ in which $\bm{x}^{(p)}_i = [\bm{v_i}; \bm{r}^{e_h}_i; \bm{r}^{e_t}_i]$ and $d_p = d_w + 2\times d_r$.

However, aside from the relative position features, we argue that the embeddings of both the head entity $e_h$ and tail entity $e_t$ are also vitally significant for relation extraction task, since the ultimate goal of this task is to predict the relationship between these two entities.
This hypothesis is further verified by our quantitative and qualitative analyses in later experiments (Section \ref{sec:ablation_study} and \ref{sec:casestudy}). 
The empirical results show that our proposed embedding can outperform the widely-used way in prior works \cite{ji2017distant}. 

In particular, we propose a novel entity-aware word embedding approach to enrich the traditional word embeddings with features of the head and tail entities. To this end, a position-wise gate mechanism is naturally leveraged to dynamically select features between relative position embedding and entity embeddings. Formally, the embeddings of head and tail entities are denoted as $\bm{v}^{(h)}$ and $\bm{v}^{(t)}$ respectively. The position-wise gating procedure is formulated as
\begin{align}
	\bm{\alpha} &= \sigmoid(\lambda \cdot (\bm{W}^{(g1)} \bm{X}^{(e)} + \bm{b}^{(g1)})), \\ \label{eq:hyperparameter}
	\tilde{\bm{X}}^{(p)} &= \tanh(\bm{W}^{(g2)}\bm{X}^{(p)} + \bm{b}^{(g2)}), \\
	\bm{X} &= \bm{\alpha} \cdot \bm{X}^{(e)} + (1-\bm{\alpha}) \cdot \tilde{\bm{X}}^{(p)},\\
	\text{where,}~\bm{X}^{(e)} &= [\bm{x}^{(e)}_i]_{i=1}^{n},~\forall \bm{x}^{(e)}_i = [\bm{v}_i; \bm{v}^{(h)}; \bm{v}^{(t)}],  
\end{align}
in which $\bm{W}^{(g1)}\in\mathbb{R}^{d_h \times 3d_w}$ and $\bm{W}^{(g2)}\in\mathbb{R}^{d_h \times d_p}$ are learnable parameters, $\lambda$ is a hyper-parameter to control smoothness, and $\bm{X} = [\bm{x}_1, \dots, \bm{x}_n] \in\mathbb R^{d_h \times n}$ containing the entity-aware embeddings of all tokens from the sentence.

\subsection{Self-Attention Enhanced Neural Network} \label{sec:self_attn_enhanced}

Previous works of relation extraction mainly employ a piecewise convolutional neural network (PCNN) \cite{zeng2015distant} to obtain contextual representation of sentences due to its capability of capturing local features, less computation and light-weight structure. However, some previous works \cite{vaswani2017attention} find that CNNs cannot reach state-of-the-art performance on a majority of natural language processing benchmarks due to a lack of measuring long-term dependency, even if stacking multiple modules. This motivates us to enhance the PCNN with another neural module, which is capable of capturing long-term or global dependencies to produce complementary and more powerful sentence representation. 

Hence, we employ a self-attention mechanism in our model due to its parallelizable computation and state-of-the-art performance.
Unlike existing approaches that sequentially stack self-attention and CNN layers in a cascade form \cite{yu2018qanet,wu2019pay}, we arrange these two modules in parallel so they can generate features describing both local and long-term relations for the same input sequence. Since each bag may contain many sentences (up to 20), a light-weight networks that can can efficiently process these sentences simultaneously is more preferable, such as PCNN that is the most popular module for relation extraction. 
For this reason, there is only one light-weight self-attention layer in our model. This is contrast to \citeauthor{yu2018qanet}~\shortcite{yu2018qanet} and \citeauthor{wu2019pay}~\shortcite{wu2019pay} who stack both modules many times repeatedly. Our experiments show that two modules arranged in parallel manner consistently outperform stacking architectures that are even equipped with additional residual connections \cite{he2016deep}). The comparative experiments will be elaborated in Section \ref{sec:main_res} and \ref{sec:ablation_study}.

\paragraph{Piecewise Convolutional Neural Network} This section provides a brief introduction to PCNN as a background for further integration with our model, and we refer readers to \citeauthor{zeng2015distant}~\shortcite{zeng2015distant} for more details. Each sentence is divided into three segments w.r.t. the head and tail entities. Compared to the typical 1D-CNN with max-pooling \cite{zeng2014relation}, piecewise pooling has the capability to capture the structure information between two entities. 
Therefore, instead of using word embeddings with relative position features $\bm{X}^{(p)}$ as the input, we here employ our entity-aware embedding $\bm{X}$ as described in Section \ref{sec:ent_aware} to enrich the input features. First, 1D-CNN is invoked over the input, which can be formally represented as
\begin{equation}
\bm{H} = \onedcnn(\bm{X}; \bm{W}^{(c)}, \bm{b}^{(c)}) \in\mathbb{R}^{d_c \times n},
\end{equation}
where, $\bm{W}^{(c)} \in\mathbb{R}^{d_c \times m \times d_h}$ is convolution kernel with window size of $m$ (i.e., $m$-gram). Then, to obtain sentence-level representation, a piecewise pooling performs over the output sequence, i.e., $\bm{H}^{(c)} = [\bm{h}_1, \dots, \bm{h}_n]$, which is formulated as
\begin{equation} \label{eq:piecewise_pool}
 \bm{s} = \tanh([\pool(\bm{H}^{(1)}); \pool(\bm{H}^{(2)}); \pool(\bm{H}^{(3)})]).
\end{equation}
In particular, $\bm{H}^{(1)}$,  $\bm{H}^{(2)}$ and $\bm{H}^{(3)}$ are three consecutive parts of $\bm{H}$, obtained by dividing $\bm{H}$ according to the positions of head and tail entities. Consequently, $\bm{s} \in\mathbb R^{3d_c}$ is the resulting sentence vector representation. 

\paragraph{Self-Attention Mechanism} To maintain efficiency of proposed approach, we adopt the recently-promoted self-attention mechanism \cite{liu2016learning,lin2017structured,shen2019tensorized,li2018hierarchical,liu2019GPN} for compressing a sequence of token representations into a sentence-level vector representation by exploiting global dependency, rather than computation-consuming pairwise ones \cite{vaswani2017attention}. It is used to measure the contribution or importance of each token to relation extraction task w.r.t. the global dependency. Formally, given the entity-aware embedding $\bm{X}$, we first calculate attention probabilities by a parameterized compatibility function, i.e., 
\begin{align}
	&\bm{A} = \bm{W}^{(a2)} \sigma(\bm{W}^{(a1)}\bm{X} + \bm{b}^{(a1)} ) + \bm{b}^{(a2)}, \\
	&\bm{P}^{(A)} = \softmax(\bm{A}),
\end{align}
where, $\bm{W}^{(a1)}, \bm{W}^{(a2)} \in\mathbb R^{d_h \times d_h}$ are learnable parameters, $\softmax(\cdot)$ is invoked over sequence, and $\bm{P}^{(A)}$ is resulting attention probability matrix. Then, the result of self-attention mechanism can be calculated as
\begin{equation}
    \bm{u} = \sum \bm{P}^{(A)} \odot \bm{X},
\end{equation}
in which, $\sum$ is performed along sequential dimension and $\odot$ stands for element-wise multiplication. And, $\bm{u} \in\mathbb R^{d_h}$ is also a sentence-level vector representation which is a complement to PCNN-resulting one, i.e., $\bm{s}$ from Eq.(\ref{eq:piecewise_pool}).

\subsection{Selective Gate} \label{sec:selective_gate}

Given a sentence bag $B = [s_1, \dots, s_m]$ with common entity pair, where $m$ is the number of sentences. As elaborated in Section \ref{sec:self_attn_enhanced}, we can obtain $\bm{S} = [\bm{s}_1, \dots, \bm{s}_m]$ and $\bm{U} = [\bm{u}_1, \dots, \bm{u}_m]$ for each sentence in the bag, which are derived from PCNN and self-attention respectively. 

Unlike previous works under multi-instance framework that frequently use a selective attention module to aggregate sentence-level representations into bag-level one, we propose a innovative selective gate mechanism to perform this aggregation. The selective gate can mitigate problems existing in distantly supervised relation extraction and achieve a satisfactory empirical effectiveness. Specifically, when handling the noisy instance problem, selective attention tries to produce a distribution over all sentence in a bag; but if there is only one sentence in the bag, even the only sentence is wrongly labeled, the selective attention mechanism will be low-effective or even completely useless. 
Note that almost $80\%$ of bags from popular relation extraction benchmark consist of only one sentence, and many of them suffer from the wrong label problem. 
In contrast, our proposed gate mechanism is competent to tackle such case by directly and dynamically aligning low gating value to the wrongly labeled instances and thus preventing noise representation being propagated. 

Particularly, a two-layer feed forward network is applied to each $\bm{u}_j$ to sentence-wisely produce gating value, which is formally denoted as 
\begin{align}
	g_j = \sigmoid(\bm{W}^{(g1)} \sigma(\bm{W}^{(g2)} \bm{u}_j &+ \bm{b}^{(g2)} ) +  \bm{b}^{(g1)}), \\
	\notag &~\forall j = 1, \dots, m, 
\end{align}
where, $\bm{W}^{(g1)} \in\mathbb R^{3d_c \times d_h}$, $\bm{W}^{(g2)} \in\mathbb R^{d_h \times d_h}$, $\sigma(\cdot)$ denotes an activation function and $g_j \in (0, 1)$. Then, given the calculated gating value, an mean aggregation performs over sentence embeddings $[\bm{s}_j]_{j=1}^m$ in the bag, and thus produces bag-level vector representation for further relation classification. This procedure is formalized as 
\begin{align} \label{eq:aggreation}
    \bm{c} = \dfrac{1}{m} \sum_{j=1}^{m} g_j \cdot \bm{s}_j
\end{align}

Finally, $\bm{c}$ is fed into a multi-layer perceptron followed with $|C|$-way $\softmax$ function (i.e., an $\mlp$ classifier) to judge the relation between head and tail entities, where $|C|$ is the number of distinctive relation categories. This can be regarded as a classification task \cite{DBLP:conf/cikm/LongCZZ12}. Formally, 
\begin{equation} \label{eq:predicted_distribution}
	\bm{p} = \softmax(\mlp(\bm{c})) \in\mathbb{R}^{|C|}.
\end{equation}

\subsection{Model Learning}

We minimize negative log-likelihood loss plus $L_2$ regularization penalty to train the model, which is written as
\begin{equation} \label{loss}
    L_{NLL} = - \dfrac{1}{|\mathcal{D}|}  \sum\nolimits_{k=1}^{|\mathcal{D}|} \log \bm{p}^k_{(i=y^k)}+\beta||\theta||^2_2
\end{equation}
where $\bm{p}^k$ is the predicted distribution from Eq.(\ref{eq:predicted_distribution}) for the $k$-th example in dataset $\mathcal{D}$ and $y^k$ is its corresponding distant supervision label. 

\section{Experiments}

\begin{table*}[t]\small
	\centering
	\begin{tabular}{lcccccccccccc}
		\toprule
		\multicolumn{1}{l}{\bf Approach }&\multicolumn{4}{c}{\bf One}&\multicolumn{4}{c}{\bf Two}&\multicolumn{4}{c}{\bf All}\\
		\midrule
		\textbf{P@N (\%)} &100&200&300&Mean&100&200&300&Mean&100&200&300&Mean\\
		\midrule
		\multicolumn{13}{l}{\textit{Comparative Approaches} } \\
		\midrule
		CNN+ATT \cite{lin2016neural} &72.0&67.0&59.5&66.2&75.5&69.0&63.3&69.3&74.3&71.5&64.5&70.1\\
		PCNN+ATT \cite{lin2016neural} &73.3&69.2&60.8&67.8&77.2&71.6&66.1&71.6&76.2&73.1&67.4&72.2\\
		PCNN+ATT+SL \cite{liu2017soft} &84.0&75.5&68.3&75.9&86.0&77.0&73.3&78.8&87.0&84.5&77.0&82.8\\
		PCNN+HATT \cite{han2018hierarchical} &84.0&76.0&69.7&76.6&85.0&76.0&72.7&77.9&88.0&79.5&75.3&80.9\\
		PCNN+BAG-ATT \cite{ye2019distant}  &86.8&77.6&73.9&79.4&91.2&79.2&75.4&81.9&91.8&84.0&78.7&84.8\\
		\midrule
		\textbf{SeG} (\textit{ours}) &\textbf{94.0}&\textbf{89.0}&\textbf{85.0}&\textbf{89.3}&\textbf{91.0}&\textbf{89.0}&\textbf{87.0}&\textbf{89.0}&\textbf{93.0}&\textbf{90.0}&\textbf{86.0}&\textbf{89.3}\\
		\midrule [0.2ex]
		\multicolumn{13}{l}{\textit{Ablations} } \\
		\midrule
		SeG w/o Ent &85.0&75.0&67.0&75.6&87.0&79.0&70.0&78.6&85.0&80.0&72.0&79.0\\
		SeG w/o Gate&87.0&85.5&82.7&85.1&89.0&87.0&84.0&86.7&90.0&88.0&85.3&87.7\\
		SeG w/o Gate w/o Self-Attn&86.0&85.0&82.0&84.3&88.0&86.0&83.0&85.7&90.0&86.5&86.0&87.5\\
		SeG w/o ALL &81.0&73.5&67.3&74.0&82.0&75.0&72.3&76.4&81.0&75.0&72.0&76.0\\ 
		\midrule[0.01ex]
		SeG+ATT w/o Gate&89.0&83.5&75.7&82.7&90.0&83.5&77.0&83.5&92.0&82.0&76.7&83.6\\
		SeG+ATT&88.0&81.0&75.0&81.3&87.0&82.5&77.0&82.2&90.0&86.5&81.0&85.8\\
		SeG w/ stack&91.0&88.0&85.0&88.0&91.0&87.0&85.0&87.7&92.0&89.5&86.0&89.1\\
		\bottomrule
	\end{tabular}
	\caption{\small Precision values for the top-100, -200 and -300 relation instances that are randomly selected in terms of one/two/all sentence(s). }
	\label{tab:topnprecision}
\end{table*}

To evaluate our proposed framework, and to compare the framework with baselines and competitive approaches, we conduct experiments on a popular benchmark dataset for distantly supervised relation extraction. 
We also conduct an ablation study to separately verify the effectiveness of each proposed component, and last, case study and error analysis are provided for an insight into our model.

\paragraph{Dataset}
In order to accurately compare the performance of our model, we adopt New York Times (NYT) dataset \cite{riedel2010modeling}, a widely-used standard benchmark for distantly supervised relation extraction in most of previous works \cite{lin2016neural,zeng2015distant,han2018hierarchical,du2018multi}, which contains 53 distinct relations including a null class \textit{NA} relation. 
This dataset generates by aligning Freebase with the New York Times (NYT) corpus automatically. In particular, NYT dataset contains 53 distinct relations including a null class \textit{NA} relation referred to as the relation of an entity pair is unavailable. 
There are 570K and 172K sentences respectively in training and test set.

\paragraph{Metrics}
Following previous works \cite{zeng2015distant,lin2016neural,han2018hierarchical,du2018multi}, we use precision-recall (PR) curves, area under curve (AUC) and top-N precision (P@N) as metrics in our experiments on the held-out test set from the NYT dataset. To directly show the perfomance on one sentence bag, we also calculate the accuracy of classification (Acc.) on non-NA sentences.  

\paragraph{Training Setup}
For a fair and rational comparison with baselines and competitive approaches, we set most of the hyper-parameters by following prior works \cite{lin2017structured,han2018hierarchical}, and also use 50D word embedding and 5D position embedding released by \cite{lin2016neural,han2018hierarchical} for  initialization, where the dimension of $d_h$ equals to 150. The filters number of CNN $d_c$ equals to 230 and the kernel size $m$ in CNN equals to $3$. In output layer, we employ dropout \cite{srivastava2014dropout} for regularization, where the drop probability is set to $0.5$. To minimize the loss function defined in Eq.\ref{loss}, we use stochastic gradient descent with initial learning rate of $0.1$, and decay the learning rate to one tenth every 100K steps.

\paragraph{Baselines and Competitive Approaches} 
We compare our proposed approach with extensive previous ones, including feature-engineering, competitive and state-of-the-art approaches, which are briefly summarized in the following.

\begin{itemize}
	\item \textbf{Mintz} \cite{mintz2009distant} is the original distantly supervised approach to solve relation extraction problems with distantly supervised data.
	\item \textbf{MultiR} \cite{hoffmann2011knowledge} is a graphical model within a multi-instance learning framework that is able to handle problems with overlapping relations.
	\item \textbf{MIML} \cite{surdeanu2012multi} is a multi-instance, multi-label learning framework that jointly models both multiple instances and multiple relations. 
	\item \textbf{PCNN+ATT} \cite{lin2016neural} employs a selective attention over multiple instances to alleviate the wrongly labeled problem, which is the principal baseline of our work. 
	\item \textbf{PCNN+ATT+SL} \cite{liu2017soft} introduces an entity-pair level denoising method, namely employing a soft label to alleviate the impact of wrongly labeled problem.
	\item \textbf{PCNN+HATT} \cite{han2018hierarchical} employs hierarchical attention to exploit correlations among relations.
	\item \textbf{PCNN+BAG-ATT} \cite{ye2019distant} uses an intra-bag to deal with the noise at sentence-level and an inter-bag attention to deal with noise at the bag-level.
\end{itemize}

\begin{figure}[t]
	\centering
	\includegraphics[scale=0.38]{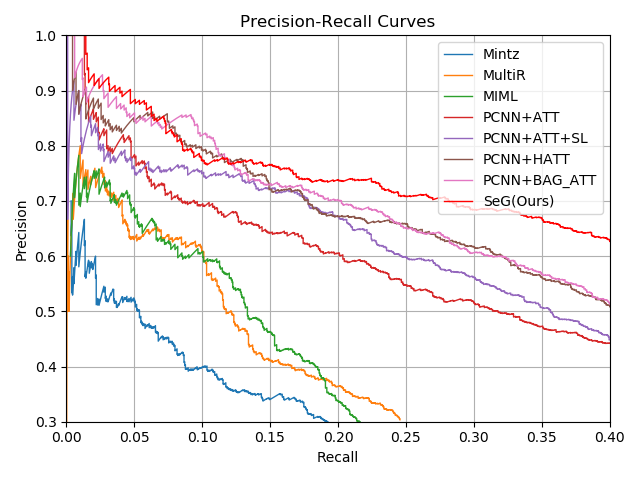}
	\caption{Performance comparison for proposed model and previous baselines in terms of precision-recall curves}
	\label{fig:prcurves}
\end{figure}

\begin{table}[t] \small
	\centering
	\begin{tabular}{lc}
		\toprule
		\textbf{Approach}&\textbf{AUC}\\
		\midrule
		PCNN+HATT& 0.42\\
		PCNN+ATT-RA+BAG-ATT& 0.42\\  
		\midrule
		\textbf{SeG} (ours)& 0.51\\  
		\bottomrule
	\end{tabular}
	\caption{\small Model comparison regarding the AUC value. The comparative results are reported by \citeauthor{han2018hierarchical}~\shortcite{han2018hierarchical} and \citeauthor{ye2019distant}~\shortcite{ye2019distant} respectively. } 
	\label{tab:aucscores}
\end{table}

\begin{table}[t] \small
	\centering
	\begin{tabular}{lcc}
		\toprule
		\textbf{Approach}&\textbf{AUC}&\textbf{Acc.}\\
		\midrule
		PCNN&0.36&83\% \\
		PCNN+ATT&0.35&78\% \\
		SeG(ours)&0.48&90\% \\
		\bottomrule
	\end{tabular}
	\caption{\small Model that is trained and tested on extracted one sentence bags from NYT dataset comparison regarding the AUC value and Acc., where Acc. is accuracy on non-NA sentences.}
	\label{tab:one-sentenceexp}
\end{table}

\subsection{Relation Extraction Performance} \label{sec:main_res}

We first compare our proposed SeG with aforementioned approaches in Table \ref{tab:topnprecision} for top-N precision (i.e., P@N). As shown in the top panel of the table, our proposed model SeG can consistently and significantly outperform baseline (i.e., PCNN+ATT) and all recently-promoted works in terms of all P@N metric. 
Compared to PCNN with selective attention (i.e., PCNN+ATT), our proposed SeG can significantly improve the performance by 23.6\% in terms of P@N mean for all sentences; even if a soft label technique is applied (i.e., PCNN+ATT+SL) to alleviate wrongly labeled problem, our performance improvement is also very significant, i.e., 7.8\%. 

Compared to previous state-of-the-art approaches (i.e., PCNN+HATT and PCNN+BAG-ATT), the proposed model can also outperform them by a large margin, i.e., 10.3\% and 5.3\% , even if they propose sophisticated techniques to handle the noisy training data. These verify the effectiveness of our approach over previous works when solving the wrongly labeled problem that frequently appears in distantly supervised relation extraction.

Moreover, for proposed approach and comparative ones, we also show AUC curves and available numerical values in Figure \ref{fig:prcurves} and Table \ref{tab:aucscores} respectively. 
The empirical results for AUC are coherent with those of P@N, which shows that, our proposed approach can significantly improve previous ones and reach a new state-of-the-art performance by handling wrongly labeled problem using context-aware selective gate mechanism. Specifically, our approach substantially improves both PCNN+HATT and PCNN+BAG-ATT by 21.4\% in aspect of AUC for precision-recall.

\subsection{Ablation Study} \label{sec:ablation_study}

\begin{table}[b] \small
	\centering
	\begin{tabular}{lc}
		\toprule
		\textbf{Approach}&\textbf{AUC}\\
		\midrule
		\textbf{SeG} (ours)& 0.51\\  
		SeG w/o Ent& 0.40\\
		SeG w/o Gate& 0.48\\
		SeG w/o Gate w/o Self-Attn& 0.47\\
		SeG w/o ALL& 0.40\\ \midrule
		SeG + ATT w/o Gate& 0.47\\
		SeG + ATT&0.47\\
		SeG w/ stack&0.48\\
		\bottomrule
    \end{tabular}
	\caption{\small Ablation study regarding precision-recall AUC value.} 
\label{tab:abl_auc_scores}
\end{table}

To further verify the effectiveness of each module in the proposed framework, we conduct an extensive ablation study in this section. 
In particular, \textit{SeG w/o Ent} denotes removing entity-aware embedding, \textit{SeG w/o Gate} denotes removing selective gate and concatenating two representations from PCNN and self-attention, \textit{SeG w/o Gate w/o Self-Attn} denotes removing self-attention enhanced selective gate. In addition, we also replace the some parts of the proposed framework with baseline module for an in-depth comparison. \textit{SeG+ATT} denotes replacing mean-pooing with selective attention, and \textit{SeG w/ stack} denotes using stacked PCNN and self-attention rather than in parallel.

\begin{figure}[t]
	\centering
	\includegraphics[scale=0.38]{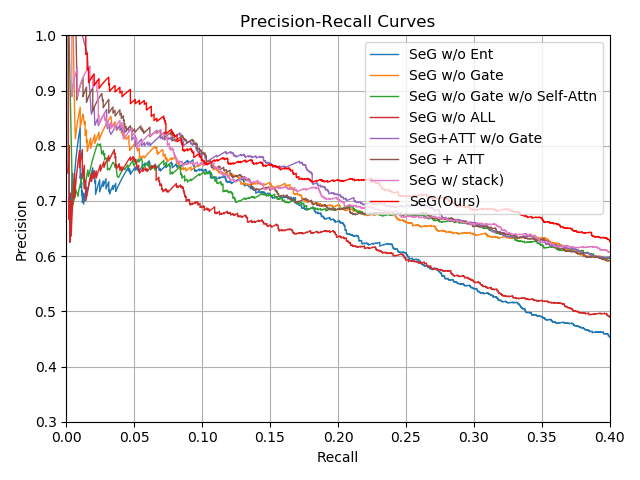}
	\caption{\small Performance comparison for ablation study under precision-recall curves}
	\label{fig:ablationpr}
\end{figure}

\begin{table*}[htbp] \small
     \centering
     \begin{tabular}{p{13pt}|p{170pt}|p{95pt}|p{50pt}|p{50pt}|p{57pt}}
     \hline
     Bag&Sentence&Relation&SeG (Ours) &SeG w/o Ent&SeG w/o GSA\\
     \hline
     B1& \textbf{Yul Kwon}, 32, of \textbf{San Mateo}, Calif., winner of last year's television contest “Survivor” and ... &\textit{/people/person/place\_lived}&Correct&Wrong&Wrong\\ 
     \hline
     B2&Other winners were Alain Mabanckou from Congo, \textbf{Nancy Huston} from \textbf{Canada} and Léonora Miano from Cameroon.&\textit{/people/person/nationality}&Correct&Correct&Wrong\\
     \hline
     \hline
     B3&... production moved to \textbf{Connecticut} to film interiors in places like Stamford, Bridgeport, Shelton, \textbf{Ridgefield} and Greenwich.&\textit{/location/location/contains}&Correct&Wrong&Correct\\ 
     \hline
     B4&... missionary \textbf{George Whitefield}, according to The Encyclopedia of \textbf{New York City}.&\textit{NA}&Correct&Wrong&Correct\\ 
     \hline
     \end{tabular}
     \caption{A case study where each bag contains one sentence. \textit{SeG w/o GSA} is an abbreviation of \textit{SeG w/o Gate w/o Self-Attn}.}
     \label{tab:casestudy}
\end{table*}

The P@N results are listed in the bottom panel of Table \ref{tab:topnprecision}, and corresponding AUC results are shown in Table \ref{tab:abl_auc_scores} and Figure \ref{fig:ablationpr}. 
According to the results, we find that our proposed modules perform substantially better than those of the baseline in terms of both metrics.
Particularly, by removing entity-aware embedding (i.e, SeG w/o Ent) and self-attention enhanced selective gate (i.e., SeG w/o Gate w/o Self-Attn), it shows 11.5\% and 1.8\% decreases respectively in terms of P@N mean for all sentences. Note that, when dropping both modules above (i.e., SeG w/o ALL), the framework will be degenerated as selective attention baseline \cite{lin2016neural}, which again demonstrates that our proposed framework is superior than the baseline by 15\% in terms of P@N mean for all sentences.  

To verify the performance of selective gate modul when handling wrongly labeled problem, we simply replace the selective gate module introduced in Eq.(\ref{eq:aggreation}) with selective attention module, namely, SeG+Attn w/o Gate, and instead of mean pooling in Eq.(\ref{eq:aggreation}), we couple selective gate with selective attention to fulfill aggregation instead mean-pooling, namely, SeG+Attn. 
Across the board, the proposed SeG still deliver the best results in terms of both metrics even if extra selective attention module is applied. 

Lastly, to explore the influence of the way to combine PCNN with self-attention mechanism, we stack them by following the previous works \cite{yu2018qanet}, i.e., SeG w/ Stack. And we observe a notable performance drop after stacking PCNN and self-attention in Table \ref{tab:abl_auc_scores}. This verifies that our model combining self-attention mechanism and PCNN in parallel can achieve a satisfactory result.

To further empirically evaluate the performance of our method in solving one-sentence bag problem, we extract only the one-sentence bags from NYT's training and test sets, which occupy ~80\% of the original dataset.
The evaluation and comparison results in Table \ref{tab:one-sentenceexp} show that compared to PCNN+ATT, the AUC improvement (+0.13) between our model and PCNN+ATT on one-sentence bags is higher than the improvement of full NYT dataset, which verifies SeG's effectiveness on one-sentence bags. 
In addition, PCNN+ATT shows a light decrease compared with PCNN, which can also support the claim that selective attention is vulnerable to one-sentence bags.

\subsection{Case Study} \label{sec:casestudy}

In this section, we conduct a case study to qualitatively analyze the effects of entity-aware embedding and self-attention enhanced selective gate. The case study of four examples is shown in Table \ref{tab:casestudy}. 

First, comparing Bag 1 and 2, we find that, without the support of the self-attention enhanced selective gate, the model will misclassify both bags into \textit{NA}, leading to a degraded performance. Further, as shown in Bag 2, even if entity-aware embedding module is absent, proposed framework merely depending on selective gate can also make a correct prediction. This finding warrants more investigation into the power of the self-attention enhanced selective gate; hence, the two error cases are shown in Bags 3 and 4.

Then, to further consider the necessity of entity-aware embedding, we show two error cases for SeG w/o Ent whose labels are \textit{/location/location/contains} and \textit{NA} respectively in Bag 3 and 4. One possible reason for the misclassification of both cases is that, due to a lack of entity-aware embedding, the remaining position features cannot provide strong information to distinguish complex context with similar relation position pattern w.r.t the two entities.

\subsection{Error Analysis}

To investigate the possible reasons for misclassification, we randomly sample 50 error examples from the test set and manually analyze them. After human evaluation, we find the errors can be roughly categorized into following two classes. 

\paragraph{Lack of background} We observe that, our approach is likely to mistakenly classify relation of almost all the sentences containing  two place entities to \textit{/location/location/contains}. However, the correct relation is \textit{/location/country/capital} or \textit{/location/country/administrative\_divisions}. This suggests that we can incorporate external knowledge to alleviate this problem possibly caused by a lack of background. 

\paragraph{Isolated Sentence in Bag} Each sentence in a bag can be regarded as independent individual and do not have any relationship with other sentences in the bag, which possibly leads to information loss among the multiple sentences in the bag when considering classification over bag level.

\section{Conclusion}

In this paper, we propose a brand-new framework for distantly supervised relation extraction, i.e., selective gate (SeG) framework, as a new alternative to previous ones. It incorporates an entity-aware embedding module and a self-attention enhanced selective gate mechanism to integrate task-specific entity information into word embedding and then generates a complementary context-enriched representation for PCNN. 
The proposed framework has certain merits over previously prevalent selective attention when handling wrongly labeled data, especially for a usual case that there are only one sentence in the most of bags. 
The experiments conduct on popular NYT dataset show that our model SeG can consistently deliver a new benchmark in state-of-the-art performance in terms of all P@N and precision-recall AUC. 
And further  ablation study and case study also demonstrate the significance of the proposed modules to handle wrongly labeled data and thus set a new state-of-the-art performance for the benchmark dataset. 
In the future, we plan to incorporate an external knowledge base into our framework, which may further boost the prediction quality by overcoming the problems with a lack of background information as discussed in our error analysis.

 \section*{Acknowledgements}    
 This research was funded by the Australian Government through the Australian Research Council (ARC) under grants LP180100654 partnership with KS computer. We also acknowledge the support of NVIDIA Corporation and Google Cloud with the donation of GPUs and computation credits respectively.

\bibliography{reference}
\bibliographystyle{aaai}

\appendix
\section{Related Work}

Recently, many works \cite{liu2017soft,du2018multi} employed selective attention \cite{lin2016neural} to alleviate wrongly labeled problem existing in distantly supervised RE. 
For example,
\citeauthor{han2018hierarchical}~\shortcite{han2018hierarchical} propose a hierarchical relation structure attention based on the insight of selective attention. 
And, \citeauthor{ye2019distant}~\shortcite{ye2019distant} extend the sentence-level selective attention to bag-level, where the bags have same relation label. 
Differing from these works suffering from one-sentence bag problem due to the defect of selective attention, our proposed approach employ a gate mechanism as an aggregator to handle this problem. 

There are several works recently proposed to couple CNN with self-attention \cite{wu2019pay,zhu2019empirical,zhou2019ssa} for either natural language processing or computer vision. For example, \citeauthor{yu2018qanet}~\shortcite{yu2018qanet} enrich CNN's representation with self-attention for machine reading comprehension.
Unlike these works stacking the two modules many times, we arrange them in parallel instead of  to ensure model's scalability. 
In addition, some previous approach explore the importance of entity embedding for relation extraction \cite{ji2017distant,beltagy2019combining}, which usually need the support external knowledge graph and learn the entity embeddings over the graph. In contrast, this approach considers the entity embeddings within a sentence and incorporate them with relative position feature without any external support. 

\end{document}